\documentclass[lettersize,journal]{IEEEtran}
\let\labelindent\relax
\usepackage{hyperref, url, amsmath, amssymb, booktabs, graphicx, bm, color, algorithm, tabularx, enumitem, multirow, threeparttable, flushend}
\usepackage[noend]{algpseudocode}
\usepackage{xcolor}
\usepackage{footnote}
\usepackage{soul}
\graphicspath{{images/}}

\newcolumntype{Y}{>{\centering\arraybackslash}X}

\makeatletter

\newcommand{\Rmnum}[1]{\expandafter\@slowromancap\romannumeral #1@}
\makeatother

\title{\LARGE \bf SkinGrip: An Adaptive Soft Robotic Manipulator with Capacitive Sensing for Whole-Limb Bed Bathing Assistance}

\author{Fukang Liu$^1$, Kavya Puthuveetil$^2$, Akhil Padmanabha$^2$, Karan Khokar$^2$, Zeynep Temel$^2$, and Zackory Erickson$^2$% <-this % stops a space
\thanks{$^1$Fukang Liu is with the Institute for Robotics and Intelligent Machines, Georgia Institute of Technology, GA, USA.
}%
\thanks{$^2$Kavya Puthuveetil, Akhil Padmanabha, Karan Khokar, Zeynep Temel, and Zackory Erickson are with the Robotics Institute, Carnegie Mellon University, PA, USA.}%
}

\begin{document}

\setlength{\textfloatsep}{10pt}
\maketitle
\thispagestyle{empty}
\pagestyle{empty}
\begin{abstract}

% Assisting individuals with disabilities and the elderly with bathing has always been a crucial healthcare task. Robotics has been increasingly recognized as a feasible solution for this task. 
% Robotics has been increasingly recognized as a feasible solution for bathing, potentially alleviating labor shortages and high costs of care and providing consistent, gentle care for individuals with physical disabilities. 
Robotics presents a promising opportunity for enhancing bathing assistance, potentially to alleviate labor shortages and reduce care costs, while offering consistent and gentle care for individuals with physical disabilities. However, ensuring flexible and efficient cleaning of the human body poses challenges as it involves direct physical contact between the human and the robot, and necessitates simple, safe, and effective control. In this paper, we introduce a soft, expandable robotic manipulator with embedded capacitive proximity sensing arrays, designed for safe and efficient bed bathing assistance. We conduct a thorough evaluation of our soft manipulator, comparing it with a baseline rigid end effector in a human study involving $12$ participants across $96$ bathing trails. Our soft manipulator achieves an an average cleaning effectiveness of $88.8\%$ on arms and $81.4\%$ on legs, far exceeding the performance of the baseline. Participant feedback further validates the manipulator's ability to maintain safety, comfort, and thorough cleaning. Video
demonstrations are available on our project website~\footnote{\url{https://sites.google.com/view/softbathing}}.
% , highlighting its potential as an impactful solution in physically assistive robotics.

\end{abstract}
% \begin{IEEEkeywords}
% Soft robotics, robot-assisted bathing, capacitive sensing, human-robot interaction, assistive robotics.
% \end{IEEEkeywords}

\section{Introduction}
\label{sec:intro}

% In recent years, there has been an increase in the population of persons with disabilities and the elderly. Many in this group struggle with performing daily activities independently due to age or disability. A global survey across 70 countries found that at least $20\%$ of people with disabilities have difficulties with self-care tasks like grooming~\cite{world2011world}. 

Recent studies by the World Health Organization (WHO) indicate that more than 100 million people aged 60 years or older need physical assistance to perform essential daily tasks~\cite{world2017integrated}. Additionally, there is a global trend of population aging, with forecasts suggesting that by 2050, the percentage of individuals over the age of 60 will surpass $22\%$ of the total world population~\cite{united2019world}. However, a growing shortage of professional human caregivers in many countries exacerbates this situation, creating financial and logistical challenges~\cite{GCoABuilding}. In response, robotic caregivers offer an opportunity to enhance the independence and quality of life for those reliant on care~\cite{bilyea2017robotic}. Notably, among activities of daily living (ADLs), bathing is one of the first tasks that adults are likely to need assistance performing~\cite{chan2012measurement, bock2012towards}.

Prior literature has introduced a number of robotic devices to assist a person in bathing their body. The majority of prior work introduces rigid flat bathing tools to contact and clean small surface regions of human limbs~\cite{erickson2019multidimensional, king2010laser, erickson2022characterizing, gu2024learning}. 
However, these devices present certain challenges: the use of rigid bathing tools that come in contact with human skin may pose safety risks during interactions with individuals who require assistance, and achieving effective cleaning of a limb's entire cylindrical surface often necessitates large and complex movements with a rigid end effector.
Soft deformable devices present a promising alternative in assistive robotics, attributed to their lightweight, safe touch, and high adaptability~\cite{majidi2014soft, manti2016soft}. 
% Huang \textit{et al.}~\cite{huang2022soft}  explored the use of soft tactile sensors for cleaning tasks on various objects, including body parts.

% ADLs, especially those performed with an assistive robotic manipulator, are challenging due to safety requirements, as the robot is in close proximity and involves close contact with the human body \cite{mohebbi2020human}. Self-care tasks such as grooming and bathing are examples of such ADLs. Unintended contact by a robot or an unexpected motion can cause severe discomfort to the human user. High-forces can cause injury in extreme scenarios, thus resulting in aggravating the health of the human user rather than improving their quality of life.

\begin{figure}[t!]
%\vspace{-0.1cm}
\centerline{\includegraphics[scale=0.2, trim={7cm 2cm 2cm 0cm}, clip]{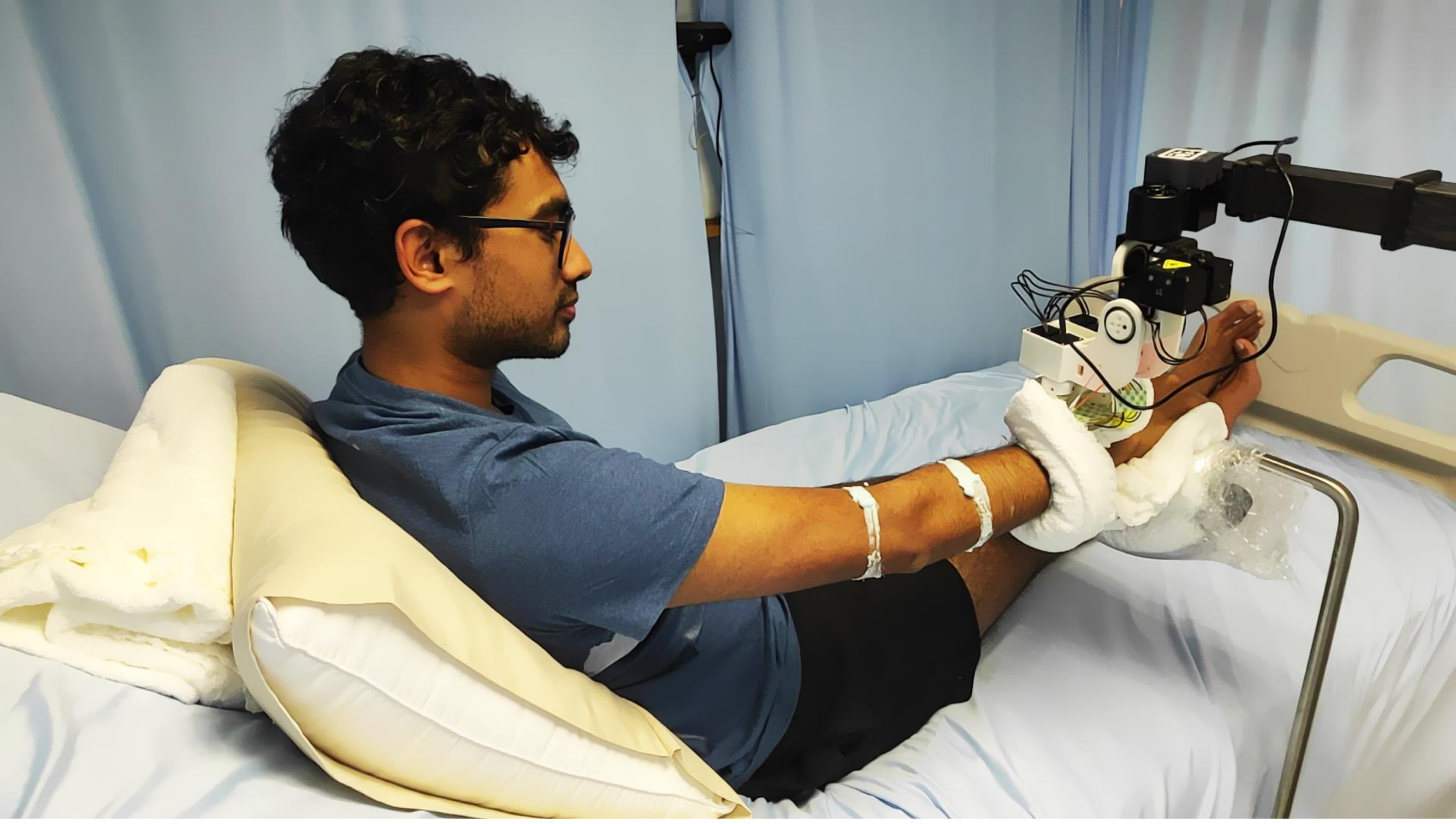}}
\caption{The soft manipulator developed in this work, mounted on the Stretch RE1 mobile manipulator, cleaning a person's right arm.
%The robot, equipped with our SkinGrip, is performing the cleaning task on a subject. Further details of the SkinGrip can be seen in Fig.~\ref{fig:3Dmodel}.
}
\label{fig:intro}
\end{figure}

In this work, we develop a robotic system to improve safety in human-robot interactions, specifically designed to aid in bed bathing tasks, as shown in Fig.~\ref{fig:intro}. This system includes a novel soft robotic manipulator—SkinGrip, equipped with two tendon-driven soft fingers. The two fingers are uniquely structured to gently encircle human limbs. Additionally, we present a capacitive servoing control scheme. This approach empowers the SkinGrip to facilitate safe and continuous contact with human limbs, when used with a Stretch RE1 mobile manipulator. Finally, we conduct a robotic bathing human study experiment to evaluate our SkinGrip against a baseline end effector~\cite{erickson2019multidimensional,erickson2022characterizing}, evaluating cleaning performance and participant perspectives, across participants with varying sized limbs. The study results demonstrate that the proposed manipulator and control strategy were able to efficiently clean the entire surface of a human limb with an average cleaning effectiveness of $88.8\%$ on arms and $81.4\%$ on legs, surpassing the baseline in both cleaning efficiency and user experience. In this work our contributions include:
\begin{enumerate}
\item We introduce a soft robotic manipulator (SkinGrip) that wraps around and conforms to human limbs for efficient bed bathing.
% \item We propose a control algorithm for the SkinGrip that ensures effective skin bathing performance and safe physical interaction. This control method integrates multidimensional capacitive sensing to maintain dynamic, full contact with the skin.
\item We implement a control approach for the SkinGrip that ensures effective skin bathing performance and safe physical interaction. This method integrates multidimensional capacitive sensing to maintain dynamic, full contact with the skin.
\item We perform a human study with 12 participants and $96$ bathing trials to assess the cleaning efficiency and obtain user feedback for both our SkinGrip and a baseline end effector. These experiments demonstrate the safety, comfort, and cleaning efficacy our proposed method.
\end{enumerate}

% A novel end effector in the form of a tendon-driven soft robotic manipulator has been designed and developed. Capacitive sensors integrated into the end effector are used to map human limb pose differential so as to enable contour following of the end effector for the bathing task. Here, bathing task refers to wiping or cleaning human limbs by a soft washcloth installed on the end effector after application of a cleaning agent such as foam or gel. The novel robotic end effector design is compared with a baseline design \cite{erickson2022characterizing} by performing bathing task on ten human subject participants, and the results are presented.

\section{Related Work}

\subsection{Robot-Assisted Bed Bathing}

% ADLs are defined as any of the basic self-care daily functions adults perform that determines their independence, such as continence, eating, grooming and walking \cite{hilgenkamp2011instrumental}. Comprehensive reviews on robotic technologies for assistance in ADLs are provided in \cite{bilyea2017robotic}, \cite{johnson2020medical} and \cite{chung2013functional}.

%Several works in literature are based on automated bathing technology \cite{bock2012towards, he2019ergonomic}, but our focus in this paper is to present a background on robot assisted bathing involving robotic manipulators. These provide contact-based solutions involving physical human-robot interaction (pHRI). Our previous work demonstrated a wearable robot that would clean a human limb by navigating over it \cite{liu2022characterization}. 
% Good cleaning performance was achieved, but the author mentions requirement of additional mechanisms to clean larger diameter limbs. 
%King et al.~\cite{king2010laser} demonstrate an end effector design for wiping a limb, where an operator selects the area to clean using a laser-interface. The main drawback is that the orientation of the end effector is always normal to the ground, which may cause discomfort to the user. Also, the system lack active sensing and compliance.

Bed bathing is a crucial aspect of nursing care, and various robot-assisted approaches have been proposed to enhance this process. Zlatintsi and Dometios et al.~\cite{zlatintsi2018multimodal,dometios2018vision,zlatintsi2020support,dometios2017real,werner2022evaluating, werner2020improving} developed a vision-based system with a soft arm for washing or wiping the back region of a person. King et al.~\cite{king2010laser} showcased an end effector designed for limb cleaning, operated via a point cloud interface for area selection. Erickson et al.~\cite{erickson2019multidimensional, erickson2022characterizing}  utilized capacitive sensing to track human motion, enabling a mobile manipulator to move a rigid end effector along a human limb. Liu et al.~\cite{liu2022characterization} introduced a mesoscale wearable robot that cleans human limbs by navigating over them. Recently, Madan et al.~\cite{madan2024rabbit} presented RABBIT, a unique robot-assisted bed bathing system that employs multimodal perception and dual compliance for a robotic manipulator to navigate a flat bathing tool along a human limb. Gu et al.~\cite{gu2024vttb} introduced a visuo-tactile Transformer-based imitation learning method (VTTB) for bathing assistance, employing multimodal sensing with a Baxter robot to track body contours and perform precise actions across different subjects. %Building on these advancements, our work introduces a soft, expandable robotic manipulator that leverages capacitive sensing control for safe and efficient bathing assistance. 

In contrast to this body of prior literature, the proposed soft end effector design in this paper is able to make contact with and bath an entire limb circumference. To date, prior literature has leveraged a planar bathing instrument that is capable of making contact with only a small region of the skin at once~\cite{erickson2019multidimensional, madan2024rabbit, gu2024vttb, gu2024learning}. In addition, we introduce a capacitive proximity sensing approach for soft robot fingers that enable maintaining full-surface contact with the human body to improve cleaning effectiveness.

% \hl{The main drawbacks of the above systems and methods were that the patient could not be bathed on the bed} \cite{zlatintsi2020support}, \hl{discomfort to user due to constrained orientation of the end-effector} \cite{king2010laser}, \hl{and that no prior research using rigid flat surface tools} \cite{erickson2019multidimensional, madan2024rabbit, gu2024vttb} \hl{has been able to clean the entire circumference of a limb.} 

% \textcolor{red}{Our work introduces a soft robotic manipulator, SkinGrip, which leverages capacitive sensing for dynamic contact maintenance and comprehensive whole-limb bathing in bed, enhancing safety, comfort, and cleaning effectiveness over previous methods.}

\vspace{-0.3cm}
\subsection{Soft Manipulators for Healthcare}

% Soft robotics has been widely used in healthcare and biomedical applications for rehabilitation and assistance purposes. \cite{chu2018soft} provides a good review of soft robotic hands used for rehabilitation. Of the 44 unique devices reviewed, 34\% of those were found to be tendon driven. Polygerinos et al.~\cite{polygerinos2015emg} present an EMG (electromyography) activated soft robotic glove that provides force to human hands to grasp objects during ADLs. \cite{cianchetti2018biomedical} provides a review on biomedical applications of soft robotics and emphasize compliant, safe and effective contact with human limbs as the advantages of the technology, especially when contact forces are high. 

% Zlatintsi \textit{et al.}~\cite{zlatintsi2020support} designed and implemented a soft robotic arm as part of an advanced bathing assistance robotic system called I-Support. The soft arm was fluid-actuated and tendon-driven. The major drawback was that the complete set-up with multiple sensors was installed in the bathroom, so the person had to physically move to the bathroom. Moreover, cleaning of upper limbs was not demonstrated.

% Huang \textit{et al.}~\cite{huang2022soft} present a soft tactile contour following robotic end effector for wiping and bathing tasks. Their systems cleans efficiently and safely due to force-sensing and depth-camera based contour adaptation. They achieve cleaning without any deep learning methods and demonstrate human limb cleaning. The major disadvantage is the bulky nature of their device as it is pneumatically powered.

Soft robotics has become increasingly prevalent in healthcare and biomedical fields, especially for rehabilitation and assistance. A comprehensive review~\cite{chu2018soft} examines soft robotic hands designed for rehabilitation, revealing that $34\%$ of the $44$ unique devices analyzed are tendon-driven. Polygerinos et al.~\cite{polygerinos2015emg} developed an electromyography (EMG)-activated soft robotic glove to aid in grasping objects for ADLs. Cianchetti et al.~\cite{cianchetti2018biomedical} provide a comprehensive overview of soft robotics in biomedical applications, highlighting compliance, safety, and effectiveness in gentle interactions with human limbs, especially under high contact forces. Furthermore, advancements in soft robotics have extended to assistive devices for bathing. Zlatintsi et al.~\cite{zlatintsi2020support} introduced a fluid and tendon-operated soft robotic arm within the I-Support bathing system. Huang et al.~\cite{huang2022soft} presented a soft pneumatic robotic tool for wiping and bathing that can adapt to slight body contours, but only makes contact with small regions of skin at a time.

In contrast to prior literature, we introduce a tendon-driven soft mechanism to achieve full limb circumference bed bathing assistance. Furthermore, we integrate capacitive proximity sensing within the soft finger architecture to sense distance from and contact with the human body. Capacitive proximity sensing enables the soft fingertips to dynamically adjust their curvature (through actuation) to ensure continuous contact and uniform pressure distribution between the finger and a human limb.

% \hl{Whereas the I-Support bathing system} \cite{zlatintsi2020support} \hl{comprised of multiple sensors installed in the bathroom, and the soft pneumatic robotic tool in} \cite{huang2022soft} \hl{is bulky, our SkinGrip system is compact in design and does not need any externally mounted sensors.} 

% \textcolor{red}{Our SkinGrip system leverages the advantages of soft robotics for safe and adaptable interactions with human limbs. \st{Unlike other systems, it integrates capacitive sensing for precise control and adaptability, dynamically maintaining contact and wrapping around the entire limb, enhancing effectiveness and safety in comprehensive bathing assistance.}}

\subsection{Capacitive Servoing and Control}

Capacitive sensing is widely utilized in healthcare for applications ranging from classifying gait phases with shoe insole sensors~\cite{park2023capacitive} to capturing clothing state as a person is getting dressed using capacitive sensors embedded in the garment~\cite{chu2019capacitiveDressing}. Similarly, Erickson et al.~\cite{erickson2019multidimensional, erickson2022characterizing} developed a capacitive servoing method using an array of capacitive sensors on a flat rigid tool for robotic bathing and dressing assistance. This approach enabled a robot end effector to traverse the contours of a human limb, but was only demonstrated to clean off the top surface of a limb. Building on these concepts, our work also employs capacitive proximity sensing, but focuses on defining control trajectories for a new manipulator designed with soft fingers that can safely encircle and clean the entire circumference of a limb, addressing the limitations of previous designs.

\vspace{-0.3cm}
\section{Design, Fabrication, and Capacitive Sensing}
\label{sec:design}

In this section, we detail the design and prototyping process of the SkinGrip, as well as a baseline end effector. Additionally, we characterize the integrated capacitive sensors as the SkinGrip interacts with human limbs. This data validation confirms that capacitance measurements can be used to detect the proximity between the SkinGrip and human skin.

\begin{figure*}[t]
\vspace{-0.1cm}
\centerline{\includegraphics[scale=0.14]{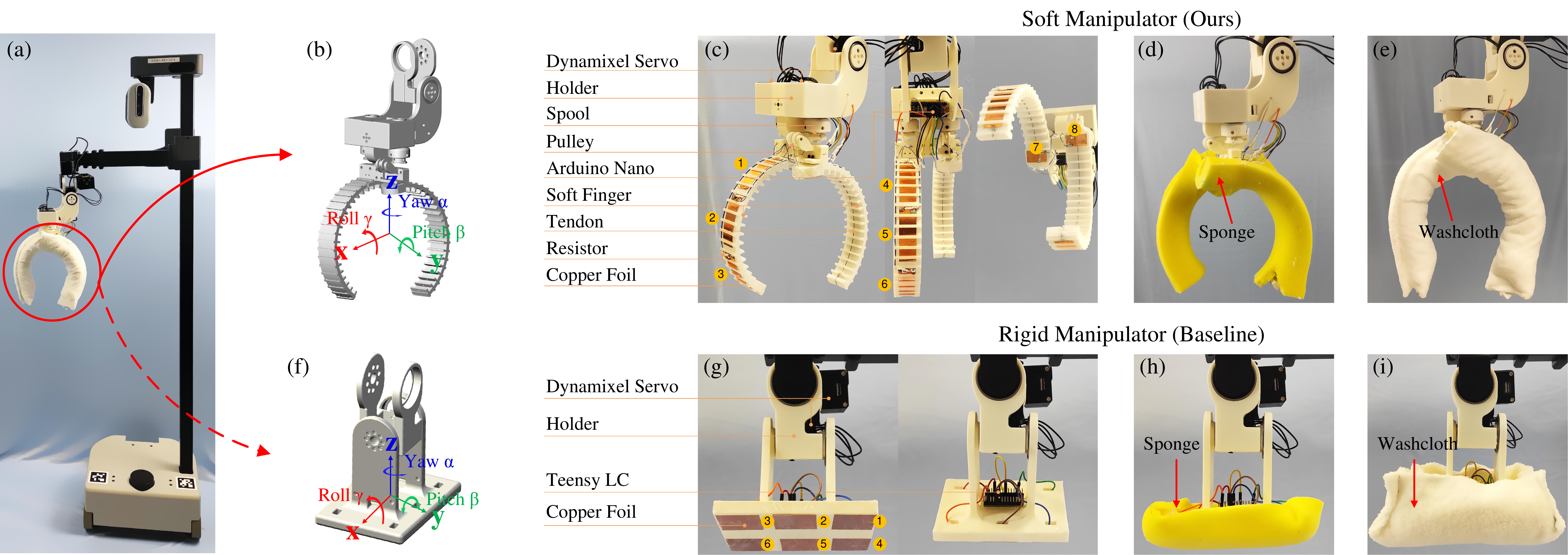}}
\caption{3D model and prototype of the proposed SkinGrip, as well as the baseline end effector. (a) The SkinGrip is mounted on a Stretch RE1 robot for cleaning tasks. (b) 3D model of the SkinGrip. (c) Real-world SkinGrip, equipped with eight copper foils labeled as: $1$-left top, $2$-left middle, $3$-left bottom, $4$-right top, $5$-right middle, $6$-right bottom, $7$-left inner, and $8$-right inner. (d) \& (e) The SkinGrip equipped with a sponge and then wrapped with a washcloth. (f) 3D model of the baseline end effector. (g) Real-world baseline end effector, equipped with six copper foils labeled as: $1$-left top, $2$-left middle, $3$-left bottom, $4$-right top, $5$-right middle, and $6$-right bottom. (h) \& (i) The baseline end effector equipped with a sponge and then wrapped with a washcloth. For both the SkinGrip and the baseline end effector, the capacitive sensors are encased in a protective plastic film to prevent direct contact with the washcloth and human skin.}
\vspace{-0.5cm}
\label{fig:3Dmodel}
\end{figure*}

% \subsection{3D Model and Prototype of the SkinGrip}
\subsection{Design, Sensing, and Actuation of the SkinGrip}
\label{sec:skingrip}

\textbf{Soft Gripper Design and Fabrication} The proposed SkinGrip design is illustrated in Fig.~\ref{fig:3Dmodel}(b)-(e). It comprises two soft fingers designed to wrap around a human limb. The size of the soft gripper was selected based on anthropometric data~\cite{tilley2001measure} to accommodate a wide range of adult limb diameters ($8$ cm to $16$ cm), ensuring it can effectively wrap around and clean the entire circumference of the limb. We optimized the width and thickness of the soft fingers to balance friction and contact area: a width that minimizes excessive friction while maintaining sufficient contact area, and a thickness that ensures flexibility without causing discomfort. Consequently, the final dimensions of the soft gripper are $160$ mm in diameter, $30$ mm in width, and $3$ mm in thickness. Each finger is tendon-actuated and powered by two Dynamixel XM430-W350-T motors. The motors bends the fingers by pulling tendons embedded within the soft material (Fig.~\ref{fig:3Dmodel}(c)), causing them to curl inward and wrap around the limb when tensioned. The required bending curvature resembles a circular arc and adapts to the shape of the limb, ensuring \textit{full contact} without causing discomfort or excessive pressure. Our approach is based on the theoretical model presented by~\cite{liu2023hybrid}, which provides the necessary equations of curvature.

\textbf{Capacitive Sensing Integration} Each finger is equipped with four capacitive proximity sensing electrodes: three are located on the back of the finger (due to the tendon routing and obstructions on the inside of the finger) with dimensions of $20\times65$ mm. Each electrode generates an electric field that permeates through the low conductivity TPU and builds charge rapidly in the presence of the high conductivity human body. And one is placed on the inner surface with dimensions $15\times25$ mm, shown in Fig.~\ref{fig:3Dmodel}(c). These electrodes are connected to an Arduino Nano microcontroller board with a 5M$\Omega$ resistor. The capacitance measured from these electrodes vary based on the distance between the electrode and an external conductive object, such as a human limb. We leverage these measurements to quantify distance and contact between human skin and a washcloth that covers the finger and capacitive electrodes (Fig.~\ref{fig:3Dmodel}(e)). Additionally, a layer of shape memory foam, Fig.~\ref{fig:3Dmodel}(d), is added between the soft finger and the washcloth to add compliance and support safe contact with human skin. Note that the capacitive sensors are encased in a protective plastic film, which prevents direct contact with both the washcloth and human skin. This design minimizes interference from mild moisture, ensuring consistent sensor performance during bathing tasks. We fabricate the soft finger and other rigid components, such as the holder, spool, and pulley (Fig.~\ref{fig:3Dmodel}(c)) of the SkinGrip using a Raise3D E2 printer with thermoplastic polyurethane (TPU) and polylactic acid (PLA), respectively.

\begin{figure*}[t]
\vspace{-0.1cm}
\centerline{\includegraphics[scale=0.155]{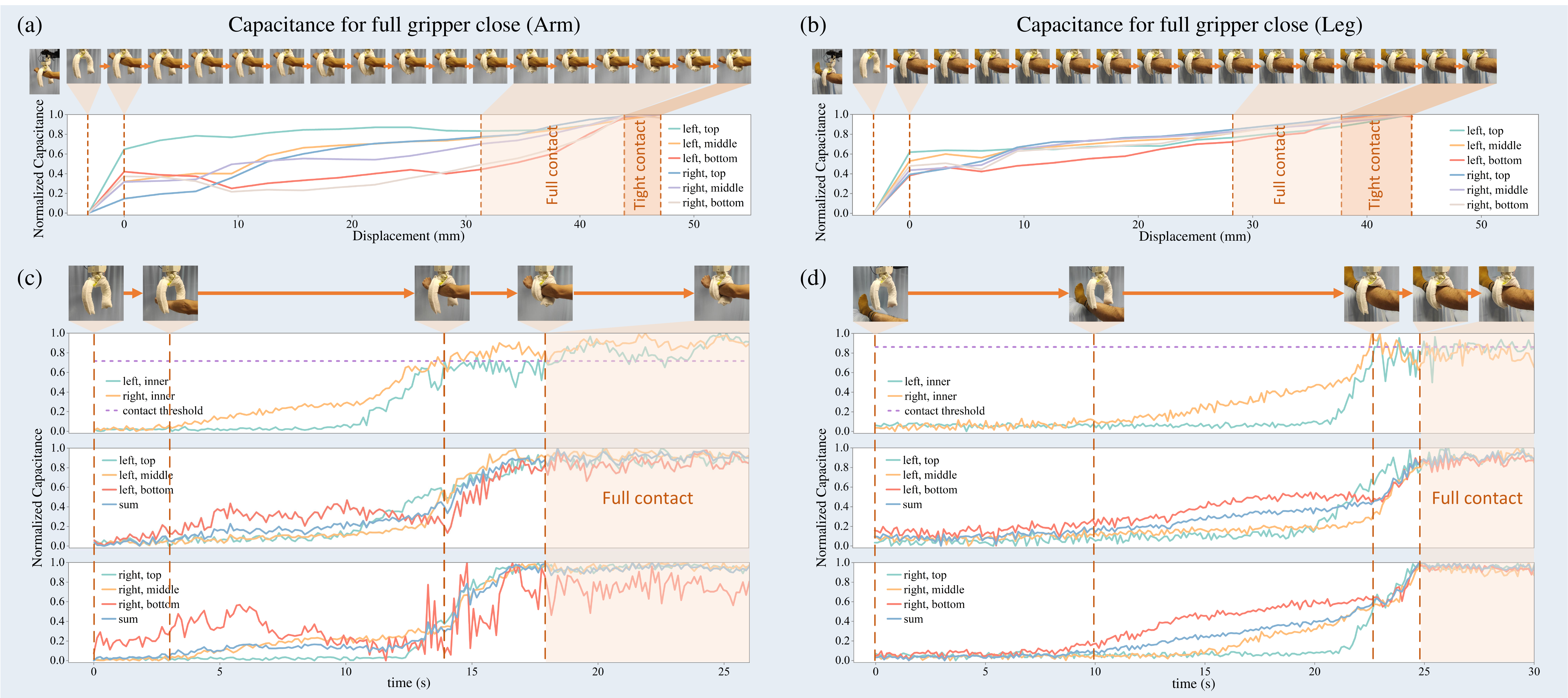}}
\caption{Capacitance measurements from all eight capacitive electrodes as the robot maneuvers the SkinGrip around an individual's limbs. (a) \& (b) Capacitance values of six capacitive electrodes during various displacements of the tendon line as the manipulator wraps around the arm and lower leg, respectively. (c) \& (d) Capacitance values of the eight capacitive electrodes as the SkinGrip approaches, wraps around, and maintains dynamic \textit{full contact} with the arm or lower leg, respectively. We normalize all signals to the range of [0, 1].}
\vspace{-0.4cm}
\label{fig:cap}
\end{figure*}

\textbf{Soft Finger Actuation and Capacitive Feedback} In order to define a control strategy for the SkinGrip, we first quantify the relationship between capacitance measurements from the embedded electrodes and displacement of the fingers around a human limb. We have the two fingers of the SkinGrip wrap around an individual's limbs, which remain stationary throughout the entire procedure. Fig.~\ref{fig:cap}(a) and (b) depict the correlation between capacitance values and tendon line displacement as soft fingers wrap around a human arm and lower leg, respectively. The results show that capacitance values increase with the tendon's displacement, indicating an increasing level of tightness between the soft finger and the limb. \textit{Full contact} refers to the condition in which the inner surface of the soft finger is in complete contact with the human skin. Furthermore, as the servos continue to apply tension on the tendon, the soft finger compresses the human skin until it achieves \textit{tight contact}, at which point the capacitance values stabilize. Notably, capacitance measurements from all sensors increase as the fingers continue to tighten around a human limb, and this trend is consistent for both arms and legs.
% Notably, there is little change in capacitance values during this phase of tight contact. Additionally, the trend in capacitance values for the human arm closely mirrors that of the human lower leg. Therefore, capacitance data can effectively quantify contact between the soft finger and the human skin. 

The robot captures capacitance measurements from each electrode at a frequency of 10~Hz. At each time step, the robot will take control actions based on a windowed observation of the most recent capacitance measurements $\bm{C}^s_{t-4:t}\in\mathbb{R}^{8\times 5}$ from the capacitive sensor from the time step $t-4$ to the current time step $t$. Subsequently, we normalize all real-time capacitance values to the range $[0,1]$ using the formula $\bm{C}_{norm}^s=\frac{\sum\bm{C}^s_{t-4:t}/5-\bm{C}_{min}^s}{\bm{C}_{max}^s-\bm{C}_{min}^s}$
% \begin{equation}
% \label{equ:getcap}
% \begin{aligned}
% \bm{C}_{norm}^s=\frac{\sum\bm{C}^s_{t-4:t}/5-\bm{C}_{min}^s}{\bm{C}_{max}^s-\bm{C}_{min}^s}
% \end{aligned}
% \end{equation}
where the superscript $s$ represents the SkinGrip. Normalization of these values, determined by the maximum $\bm{C}_{max}^s$ and minimum $\bm{C}_{min}^s\in\mathbb{R}^8$ capacitance values, ensures consistent sensitivity across different conditions. We establish a threshold $\bm{C}_{th}^{s1}\in\mathbb{R}^6$ for six of the eight electrodes, specifically: $1$-left top, $2$-left middle, $3$-left bottom, $4$-right top, $5$-right middle, and $6$-right bottom (Fig.~\ref{fig:3Dmodel}(c)), to ensure \textit{full contact} between the soft fingers and the human skin. Here, the superscript `$s1$' signifies thresholds related to these six electrodes. If the capacitance falls below the threshold, indicating insufficient contact, the control system adjusts the finger's position to re-establish \textit{full contact}. 
%This approach ensures effective cleaning by the SkinGrip and avoids applying excessive pressure or \textit{tight contact} that could lead to discomfort. 
The regulation of the two Dynamixel motors, which operate the two soft fingers via tendons, is defined as: 
\vspace{-0.1cm}
\begin{equation}
\label{equ:motor_step}
\begin{aligned}
\bm{\theta}_t = \bm{\theta}_{t-1} + \bm{K}_p^{\theta}\bm{e}_s
\end{aligned}
\end{equation}
where $t$ is the time step; $\bm{\theta}\in\mathbb{R}^2$ is the position of the two motors; $\bm{e}_s = \bm{C}_{th}^{s1} - (\bm{C}_{norm}^s)_{(1-6)} \in\mathbb{R}^{6}$ is a vector of errors between the current capacitance and the desired values (i.e., six thresholds)\footnote{In the subscript notation used within these formulas, numerical values directly following the electrode notation refer to the specific electrode number. Thus, $(\bm{C}_{norm}^s)_{(1-6)}$ refers to the values from the 1st to the 6th electrodes, inclusive. Similarly, $({C}_{norm}^s)_{7}$ corresponds to the value for the 7th electrode.}; and $\bm{K}_p^{\theta} \in\mathbb{R}^{2\times 6}$ is a gain matrix.

\textbf{Baseline Rigid End Effector Design} We contrast the SkinGrip to a baseline rigid flat end effector based on prior work~\cite{erickson2019multidimensional,erickson2022characterizing}, shown in Fig.~\ref{fig:3Dmodel}(f)-(i). This baseline end effector consists of six capacitive electrodes arranged in a $2\times3$ grid, adhered to the bottom surface of the rigid tool, as shown in Fig.~\ref{fig:3Dmodel}(g). These copper foils are connected to a Teensy LC microcontroller board. The flat tool is also covered with shape memory foam and a washcloth, shown in Fig.~\ref{fig:3Dmodel}(h)-(i). This rigid flat tool was manufactured using a Raise3D E2 printer with PLA material. Finally, we mount the two end effectors to a Stretch RE1 mobile manipulator to perform body bathing along the human limbs, as shown in Fig.~\ref{fig:3Dmodel}(a).

\vspace{-0.3cm}
% \subsection{End Effector Control from Capacitive Sensing}
\subsection{End Effector Operation from Capacitive Sensing}
\label{sec:Multidimensional-Capacitive-Sensing}

% Our goal is to instruct the robot to maneuver both the SkinGrip and the baseline end effector along a human limb to provide bathing assistance. The Stretch RE1 must control the end effector's position and orientation relative to a person's limb, using capacitance measurements obtained from the integrated capacitive proximity sensors.
% In this section, we discuss the capacitive servoing control scheme employed to maneuver the Stretch RE1 end effector along a human limb.

Our goal is to instruct the Stretch RE1 to maneuver both the SkinGrip and the baseline end effector along a human limb for bathing assistance. The robot adjusts the end effector's position and orientation using capacitance measurements from the integrated proximity sensors, enabling adaptive interaction with the human limb.

\textbf{SkinGrip Operation} As mentioned in Section~\ref{sec:skingrip}, we define a threshold $\bm{C}_{th}^{s1}\in\mathbb{R}^6$ to ensure \textit{full contact} between the two soft fingers and the human skin. Additionally, we require a threshold ${C}_{th}^{s2}\in\mathbb{R}$ for the $7$-left inner electrode. This is crucial to detecting contact along the approach axis between the upper part of the soft finger and human skin. Leveraging the left and right inner electrodes (Fig.~\ref{fig:3Dmodel} (c)), the two-electrode capacitive sensor is capable of measuring roll rotation $\gamma$, Fig.~\ref{fig:3Dmodel} (b), on the limb's  surface. Note that our system does not infer translations along the $x$- and $y$-axes or rotations ($\alpha$ and $\beta$) around the $z$-axis and $y$-axis (Fig.~\ref{fig:3Dmodel} (b)). Instead, the robot translates its end effector at a constant velocity of 4 cm/s along the central axis of a human limb to ensure bathing task progress. 
% we have pre-set fixed values for the robot's movement along the $y$-axis and the rotation angle for the SkinGrip's pitch movement around the $y$-axis [see more details in \ref{sec:study}]. % However, we will manually control the robot's movement along the $y$-axis and enable the SkinGrip to rotate in the pitch direction around the $y$-axis is not sensor controlled. [see more details in \ref{sec:study}].

\begin{algorithm}[t]\small
\caption{SkinGrip Control} 
\label{alg:control}
\begin{algorithmic}[1]
\State{{\bf Given:} $\bm{C}_{min}^s,\ \bm{C}_{max}^s,\ \bm{C}_{th}^s=[\ {C}_{th}^{s1},\ {C}_{th}^{s2}, \ {C}_{th}^{s3}\ ],\ \bm{K}_p^s$, $\Delta_s=3 \ \text{mm}$,\ $\eta_s=0.01$}
\State{Initialize: $t\gets 1$, $z_t$, $\gamma_t$, Phase $\gets Approach$}
\State{Collect initial capacitance data $\bm{C}_{t}^s$, for $t=1$ to $t=4$}
\While{\textit{task not completed}}
    \State{$\bm{C}_{norm}^s\gets$ Get normalized capacitance}
    \State{Compute $z_s,\ \gamma_s$ using (\ref{equ:soft_step})}
    \If{ Phase $== Approach$}
        \While{$(\bm{C}_{norm}^s)_7<\bm{C}_{th}^{s2}$}
            \State{$z_t\gets z_{t-1}-\Delta_s$}
        \EndWhile
        \State{Phase $\gets Closure$}
    \EndIf
    \If{ Phase $== Closure$}
        \While{$\|\bm{e}_s\|<\eta_s$}
            \State{Actuate soft fingers using (\ref{equ:motor_step})}
        \EndWhile
        \State{Phase $\gets Cleaning$}
    \EndIf
     \If{ Phase $== Cleaning$}
        % \State{$\gamma\gets \gamma+ \delta \gamma$, $ z\gets z+ \delta z$}
        \State{Update $z_t$, $\gamma_t$ using (\ref{equ:soft_step})}
        \State{Actuate soft fingers using (\ref{equ:motor_step})}
        \State{Translate end effector at $4$ cm/s along central axis of limb}
    \EndIf
    \State{$t\gets t+1$}
\EndWhile
\end{algorithmic}
\end{algorithm}

Algorithm~\ref{alg:control} details the capacitive servoing control for the SkinGrip. The control for the SkinGrip is defined as:
\begin{equation}
\label{equ:soft_step}
\begin{aligned}
% & \begin{bmatrix} \delta z, \ \delta \gamma\end{bmatrix}^{T} = \bm{K}_p^s\begin{bmatrix} z_e, \ \gamma_e
& \begin{bmatrix} z_t, \ \gamma_t\end{bmatrix}^{T} = \begin{bmatrix} z_{t-1}, \ \gamma_{t-1}\end{bmatrix}^{T} + \bm{K}_p^s\begin{bmatrix} z_s, \ \gamma_s\end{bmatrix}^{T}
\end{aligned}
% \vspace{-0.1cm}
\end{equation}
where $t$ is the time step, $z$ represents the \textit{z}-coordinate value of the displacement in the SkinGrip frame, and $\gamma$ denotes the orientation values for roll displacement in the same frame (Fig.~\ref{fig:3Dmodel} (b)). The term $\bm{K}_p^s\in\mathbb{R}^{2\times 2}$ denotes a diagonal matrix for the proportional (P) gains. Specifically, $z_s= -(\bm{C}_{norm}^s)_7-C_{th}^{s2}$, and $\gamma_s= [(\bm{C}_{norm}^s)_7-C_{th}^{s2}]-[(\bm{C}_{norm}^s)_8-C_{th}^{s3}]$, where $C_{th}^{s3}\in\mathbb{R}$ is the threshold for the right inner electrode shown in Fig.~\ref{fig:3Dmodel} (c). $z_s\in\mathbb{R}$ indicates the vertical discrepancy from the target position, and $\gamma_s\in\mathbb{R}$ measures the difference in the desired rotational alignment, based on sensor readings and target thresholds. These discrepancies guide the adjustments needed to achieve and maintain the manipulator's intended position and orientation.

Next, we carried out tests to monitor the variations in capacitance values as the SkinGrip approaches a human limb and encloses the limb until \textit{full contact} is established between the two soft fingers and the human skin. This process is illustrated in Fig.~\ref{fig:cap} (c) and (d). The end effector continues to approach the human limb with its fingertips open so long as the value of the $7$-left inner electrode (Fig.~\ref{fig:3Dmodel} (c)) falls below the threshold $C_{th}^{s2}$, indicating a lack of contact between the upper portion of the soft finger and the human skin. The control system adjusts the manipulator's position $z_t$ in (\ref{equ:soft_step}) to regulate the touch interaction with the skin.

When the value of the left inner sensor reaches the threshold $C_{th}^{s2}$, the two soft fingers commence a closure phase. During the closure phase the two fingers progressively close and surround the human limb for \textit{full contact}. The closure phase ends when the sum of errors between the current and desired values of the six electrodes is minimized, indicating \textit{full contact}. This is achieved by dynamically adjusting the motor to bend the fingers. The system then maintains this \textit{full contact} dynamically, using a proportional controller (see \ref{equ:motor_step}) for continuous feedback and adjustment, and continuously adapting $z_t$ and $\gamma_t$ according with the P controller in (\ref{equ:soft_step}). This ensures the fingers adapt to changes in limb position or size, maintaining contact by adjusting tendon tension based on feedback from all electrodes. Finally, the Stretch RE1 maneuvers the soft manipulator along the central axis of the human limb at a constant velocity of $4$ cm/s, starting from one end of the limb to the other end. In this process, the manipulator's $z$ and $\gamma$ positions will be updated using \ref{equ:soft_step}, and the two soft fingers will be dynamically controlled using \ref{equ:motor_step}.

% \textbf{Baseline Rigid End Effector Control} 
\textbf{Baseline Rigid End Effector Operation}
Similarly, for the baseline end effector, we employ a $2\times3$ grid of electrodes, shown in Fig.~\ref{fig:3Dmodel} (g), to detect the relative pose of the end effector in relation to the human limb. This end effector follows the same design and control paradigm specified in~\cite{erickson2019multidimensional, erickson2022characterizing}. Specifically, the $2$-left-middle and $5$-right-middle electrodes are essential for identifying contact (along the $z$ axis) between the baseline end effector and human skin. The four electrodes, numbered $1,4,3$ and $6$, enable the detection of displacement along the $x$ axis and the measurement of the yaw rotation $\alpha$ on the surface of the limb, ensuring that the center line of the end effector remains aligned with the central axis of a human limb.
% $y$ axis. 
Furthermore, the two rows of six electrodes collectively facilitate the measurement of roll rotation $\gamma$ on the limb's surface, ensuring that the bottom surface of the end effector remains parallel to human skin. Note that we do not detect translations along the $y$ axis or rotation $\beta$ around the $y$-axis, as shown in Fig.~\ref{fig:3Dmodel} (f). The end effector translates at a fixed velocity of 4 cm/s along the axis of the limb to ensure bathing task progress.
\begin{algorithm}[t]\small
\caption{Baseline End Effector Control} 
\label{alg:baseline}
\begin{algorithmic}[1]
\State{{\bf Given:} $\bm{C}_{min}^r,\ \bm{C}_{max}^r,\ \bm{C}_{th}^r,\ \bm{K}_p^r$,\ $\Delta_r=3 \ \text{mm}$}
\State{Initialize: $t\gets 1$, $x_t$, $z_t$, $\alpha_t$, $\gamma_t$, Phase $\gets Approach$}
\State{Collect initial capacitance data $\bm{C}_{t}^r$ for $t=1$ to $t=4$}
\While{\textit{task not completed}}
    \State{$\bm{C}_{norm}^r\gets$ Get normalized capacitance}
    \State{Compute $x_r, \ z_r, \ \alpha_r, \ \gamma_r$ using (\ref{equ:rigid_step})}
    \If{ Phase $== Approach$}
        \While{all($(\bm{C}_{norm}^s)_i<(\bm{C}_{th}^r)_i$ for i in range(6))}
            \State{$z_t\gets z_{t-1}-\Delta_r$}
        \EndWhile
        \State{Phase $\gets Cleaning$}
    \EndIf
     \If{ Phase $== Cleaning$}
        \State{Update $x_t$, $z_t$, $\alpha_t$, $\gamma_t$ using (\ref{equ:rigid_step})}
        \State{Translate end effector at $4$ cm/s along central axis of limb}
    \EndIf
    \State{$t\gets t+1$}
\EndWhile

\end{algorithmic}
\end{algorithm}

Algorithm~\ref{alg:baseline} defines the capacitive servoing control strategy for the baseline rigid end effector. Similar to the approach used with the SkinGrip, at each time step, we normalize all real-time capacitance values to the range $[0,1]$ using the formula $\bm{C}_{norm}^r=\frac{\sum\bm{C}^r_{t-4:t}/n-\bm{C}_{min}^r}{\bm{C}_{max}^r-\bm{C}_{min}^r}$, where $\bm{C}_{max}^r$ and $\bm{C}_{min}^r\in\mathbb{R}^6$ denote the maximum and minimum capacitance values, respectively, for each electrode. We establish a threshold $\bm{C}_{th}^{r}\in\mathbb{R}^6$ to ensure \textit{full contact} between the baseline end effector and the human skin. The control for the baseline end effector is then defined as:
% \vspace{-0.2cm}
\begin{equation}
\label{equ:rigid_step}
\begin{aligned}
% & \begin{bmatrix} \delta x, \ \delta z, \ \delta \alpha, \ \delta \gamma\end{bmatrix}^{T} = \bm{K}_p^r\begin{bmatrix} x_e, \ z_e, \ \alpha_e, \ \gamma_e \end{bmatrix}^{T}
\begin{bmatrix} x_t, \ z_t,\ \alpha_t, \ \gamma_t\end{bmatrix}^{T} = \;& \begin{bmatrix} x_{t-1},\ z_{t-1},\ \alpha_{t-1},\ \gamma_{t-1}\end{bmatrix}^{T}\\
&+\bm{K}_p^r\begin{bmatrix} x_r, \ z_r, \ \alpha_r, \ \gamma_r \end{bmatrix}^{T}
\end{aligned}
% \vspace{-0.1cm}
\end{equation}
where $x$ and $z$ represent the \textit{x}- and \textit{z}-coordinate values of the displacements in the baseline end effector frame, and $\alpha$ and $\gamma$ denote the orientation values for roll and yaw displacements in the same frame  (Fig.~\ref{fig:3Dmodel} (f)). The term $\bm{K}_p^r\in\mathbb{R}^{4\times 4}$ denotes a diagonal matrix for the proportional (P) gains. Specifically, $x_r= (\bm{C}_{norm}^r)_1-(\bm{C}_{norm}^r)_3+(\bm{C}_{norm}^r)_4-(\bm{C}_{norm}^r)_6$ is determined by the normalized capacitance values from electrodes in both the first row (electrodes $1$ and $3$) and the second row (electrodes $4$ and $6$), which are used to measure lateral displacement along the x-axis (Fig.~\ref{fig:3Dmodel} (g)). Similarly, $z_r= (\bm{C}_{th}^{r})_2-(\bm{C}_{norm}^r)_2 +(\bm{C}_{th}^{r})_5-(\bm{C}_{norm}^r)_5$ is calculated from the difference between the threshold and normalized capacitance values of electrodes $2$ and $5$, enabling the assessment of vertical distance along the $z$-axis. The orientation angles $\alpha_r=(\bm{C}_{norm}^r)_4-(\bm{C}_{norm}^r)_1+(\bm{C}_{norm}^r)_3-(\bm{C}_{norm}^r)_6$ and $\gamma_r=(\bm{C}_{norm}^r)_1+(\bm{C}_{norm}^r)_2+(\bm{C}_{norm}^r)_3-(\bm{C}_{norm}^r)_4-(\bm{C}_{norm}^r)_5-(\bm{C}_{norm}^r)_6$ are obtained from the configuration of six electrodes in two rows, which facilitate the detection of roll and yaw movements relative to the limb's surface and its central axis, respectively. 
As with the SkinGrip, the rigid end effector follows a linear trajectory along the limb.
%The Stretch RE1 maneuvers the rigid end effector along the central axis of the human limb also at a constant velocity of $4$ cm/s, starting from one end of the limb to the other end. 
In this process, the manipulator's $x$, $z$, $\alpha$ and $\gamma$ positions will be updated using \ref{equ:rigid_step}.
% \vspace{-3mm}

% \begin{figure}
%     \centering
%     \includegraphics[width=0.75\linewidth, trim={13cm 8cm 8cm 7cm}, clip]{characterization.pdf}
%     % \vspace{-0.2cm}
%     \caption{We measure the relative contact pressure applied by the SkinGrip to a pressure-sensing mat wrapped around the forearm and lower leg. We then map the recorded pressure to each of the soft fingers. Light and dark regions of the heatmap correspond to high and low pressure respectively. The darkest (zero) regions do not indicate that there is no contact. }
%     \label{fig:characterization}
%     % \vspace{-0.5cm}
% \end{figure}

\begin{figure*}[t]
%\vspace{-0.1cm}
\centerline{\includegraphics[scale=0.27]{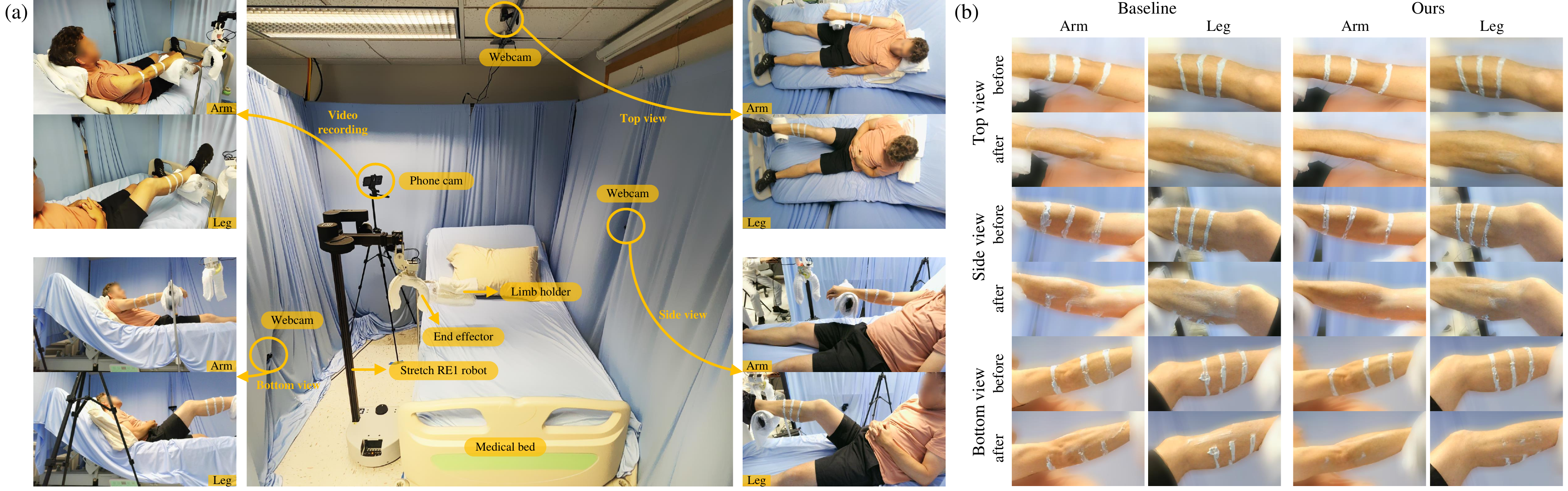}}
\caption{Real-world human study setup and cleaning performance of the two end effectors. (a) The system comprises a medical bed for participants to lie on, a Stretch RE1 robot equipped with either the soft or baseline end effector for cleaning the human limb, three webcams positioned to capture the limb from top, side, and bottom perspectives, and a phone camera for video recording. (b) Demonstrations of the cleaning capabilities of both the baseline end effector and our SkinGrip, shown in top, side, and bottom views before and after the cleaning process.}
\label{fig:setup}
\vspace{-0.5cm}
\end{figure*}

\section{Study Design}
\label{sec:study}

To assess the cleaning performance of both our SkinGrip and the baseline end effector, we conducted a human-robot study involving $12$ able-bodied adult participants [$2$ females, $10$ males, age: $26.9 \pm 6.5$ years, weight: $73.3 \pm 9.3$ kg, height: $176.6 \pm 8.2$ cm]. This study was approved by Carnegie Mellon University's Institutional Review Board under 2022.00000273.  

The experimental setup is illustrated in Fig.~\ref{fig:setup} (a). Participants lie on a hospital bed with the head and foot of the bed adjustable for comfort based on individual preferences. 
% A Stretch RE1 robot, equipped with either the soft or baseline end effector, is employed for cleaning the human limb. 
A limb holder secures the participant's limb parallel to the ground plane during the cleaning task, and three cameras are positioned to capture top, side, and bottom views of the participant's limbs. Prior to cleaning, three rings of shaving cream are applied to the limb that the robot must clean off to demonstrate skin coverage and bed bathing performance. Shaving cream was selected as it is skin-safe, straightforward to apply, and has a color that contrasts with skin tones to support visual analysis in cleaning performance. Recorded images from all three cameras are used to evaluate the cleaning performance of both the soft and baseline end effectors. An additional phone camera records video throughout the experiment. For each participant and each end effector, the robot conducted 2 bathing attempts for both the arm and the leg, resulting in 8 bathing trials per participant. We randomized ordering between the two end effectors for each participant.

% We conducted a series of $12$ evaluation sets with each end effector. Each set consisted of tests on both the arm and the lower leg, repeated twice. We randomized ordering between the two end effectors for each participant. Specifically, before the study, each participant lied on the medical bed with their arm or leg holding a pose in preparation for bathing. The limb was supported by a holder to remain stationary and parallel to the ground. We measured each participant's arm and lower leg length, as well as the diameters of five sections on each limb. These sections were equally spaced, each representing $1/5^{th}$ of the total length of the limb. 

At the beginning of an experiment with each participant, we established specific capacitance thresholds for both end effectors. The thresholds for the SkinGrip were identified at the point where participants reported comfort and confirmed \textit{full contact} between the soft finger of the end effector and their limb as the soft fingers encircled their limb. In contrast, thresholds for the rigid end effector were set based on participant reports of comfortable pressure when the end effector, held at a neutral orientation with no roll, pitch, or yaw, was placed centrally against the skin of their limb. Note that SkinGrip is designed for bed bathing rather than fully submerged or high-humidity environments such as showering. The capacitive sensors operate reliably in mild moisture conditions and are protected with a plastic film to prevent direct contact with water, the washcloth, or human skin. As a result, the system does not require full waterproofing to perform effective cleaning in bed bathing scenarios.

Prior to each bathing trial, we applied three rings of shaving cream to the three middle sections of the participant's limb, as shown in Fig.~\ref{fig:setup}. The mobile manipulator then controlled its end effector along the surface of the participant's limb, from the wrist to the shoulder for the arm and from ankle to knee for the leg (trajectory $\#1$). Once the end effector reached the end of the first trajectory, the mobile manipulator rotated its end effector by $60^{\circ}$ around the circumference of the limb (i.e., $\beta=60^{\circ}$ in Fig.~\ref{fig:3Dmodel} (b) and (f)) and then translated back to its starting position (trajectory $\#2$). This specific degree of rotation was chosen because larger rotations proved to be too challenging to execute while maintaining accuracy and comfort. For both the soft and rigid end effectors, this rotation around the limb circumference added variation between consecutive trajectories and increased surface coverage for the rigid end effector. The proposed SkinGrip uses only two trajectories to clean a limb; however, we determine that the baseline rigid end effector requires a third trajectory to increase surface coverage due to its limited non-deformable contact region between the rigid tool and a limb. For its third trajectory, the mobile manipulator rotates the baseline end effector to the opposite side of the limb circumference (i.e., $\beta=-60^{\circ}$) and then performs a bathing trajectory, translating back to the shoulder or knee (trajectory $\#3$).
% ~\textcolor{red}{This rotation was not required for the SkinGrip due to its material flexibility, which allowed it to conform to the limb's shape and ensure complete coverage in just two movements.}~\fukang{I'm uncertain if this explanation for using two movements with the SkinGrip is sufficient, especially for the leg. This method, although effective for the arm, may not guarantee complete coverage on the leg due to its larger size.} This meant that a single trial involved three movements (wrist to shoulder to wrist to shoulder) for the baseline end effector, but only two movements (wrist to shoulder to wrist) for the soft end effector. The RE1 robot operated consistently at a base translation speed of $0.04~\text{m/s}$. 

We captured images of the limb from three different views before and after cleaning. The washcloth was replaced after each trial to guarantee consistent starting conditions for every test. Participants were also asked to complete a questionnaire to evaluate the ease of use, efficiency, comfort, and preference of both the SkinGrip and the baseline end effector. Further details are available in Section~\ref{sec:response}. 

\section{Results and Discussion}

\begin{figure}[t]
%\vspace{-0.1cm}
\centerline{\includegraphics[scale=0.23]{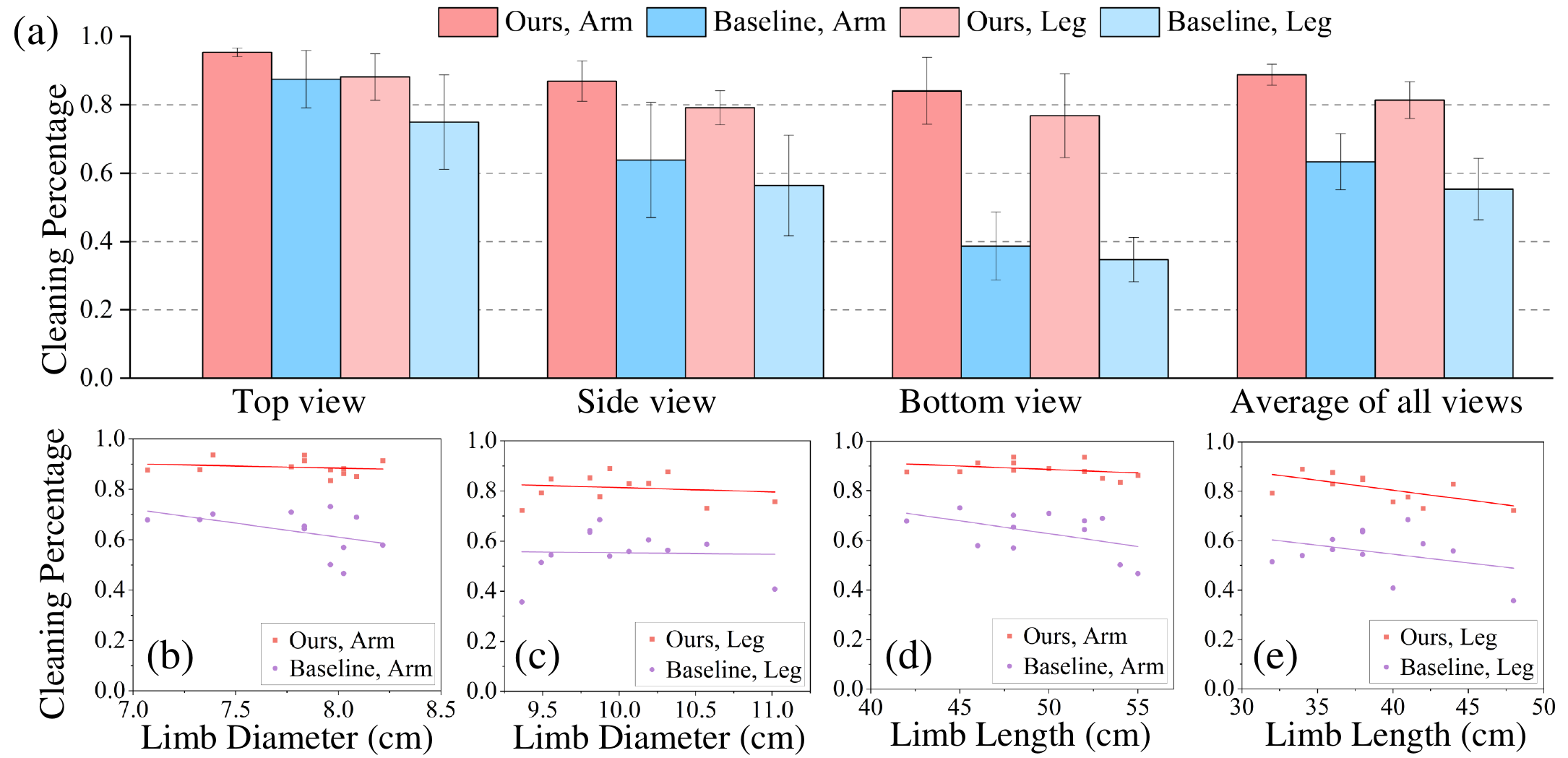}}
\caption{Evaluation of cleaning performance using the SkinGrip and baseline end effector: (a) Cleaning percentage as observed from three different views of the arm and leg. (b) and (c) Impact of limb diameter and (d) and (e) limb length on cleaning percentage, evaluated separately for the arm (b, d) and leg (c, e).}
\label{fig:percentage}
\vspace{-0.3cm}
\end{figure}

\subsection{Cleaning Performance Evaluation}

A representative illustration of the cleaning capabilities of both the soft and the baseline end effectors is presented in Fig.~\ref{fig:setup} (b). In general, we find that while the baseline end effector effectively removes most of the shaving cream from the top surface of the limb, its performance is less efficient on the sides and bottom due to its limited coverage area. To effectively clean the sides and bottom, the end effector would need to perform a full $180^\circ$ rotation and several more cleaning trajectories, well beyond the 2 trajectories used for the SkinGrip. Conversely, the SkinGrip, with only two trajectories, demonstrates visually improved cleaning performance. For the representative trial shown in Fig.~\ref{fig:setup} (b), the SkinGrip successfully removes all shaving cream from the arm and shows significantly improved performance on the leg compared to the baseline end effector.

Additionally, Fig.~\ref{fig:percentage} (a) displays the average cleaning percentages across all $12$ participants, calculated as the ratio of the area cleaned ($A_b-A_a$) over the initial area ($A_b$), where $A_b$ and $A_a$ are the areas of shaving cream before and after cleaning. These areas were determined through manual labeling. Notably, the SkinGrip (Ours) presents a consistent quantitative bathing performance increase over the current rigid manipulator (Baseline). The SkinGrip achieves an average skin bathing performance of $88.8\%$ across all trials for cleaning a participant's arm. Similarly, the SkinGrip demonstrates an average performance of $81.4\%$ across all trials for cleaning a participant's leg. 
% It is noticeable that, despite there is a slight downward trend in cleaning efficiency from the top to the side and then to the bottom view, the SkinGrip consistently exhibits robust performance. It achieves an average cleaning percentage above $84\%$ across all three views of the arm, averaging $88.8\%$ overall. Similarly, for the leg, it maintains more than $76\%$ in all three views, with an average of $81.4\%$. 
In contrast, the baseline end effector shows a pronounced decrease in cleaning effectiveness for the sides and bottom of a limb. Notably, the rigid end effector achieves an average cleaning performance of only $38.7\%$ and $34.7\%$ for the bottom of the arm and leg, respectively. The overall cleaning performance of the baseline end effector across the entire surface of a limb (all three views) averages to $63.4\%$ for the arm and $55.4\%$ for the leg across all participants.

Analysis of the scatter plots in Fig.~\ref{fig:percentage} (b)-(e) reveals distinct trends in cleaning performance between the SkinGrip and the baseline end effector across various limb dimensions. When evaluated across varying diameters of arms and legs (Fig.~\ref{fig:percentage} (b) and (c)), the SkinGrip maintains a high cleaning efficiency, irrespective of limb diameter, suggested by best-fit trend lines. In contrast, the baseline end effector demonstrates a decrease in bathing performance as the arm diameter increases, indicating a potential limitation in adapting to varying sizes. When assessing the impact of limb length (Fig.~\ref{fig:percentage} (d) and (e)) on performance, % the SkinGrip again shows a consistent cleaning percentage across the range of arm lengths and exhibits a slight decline as leg length increases, while the baseline exhibits a slight decline as both of the arm and leg length increases.
the SkinGrip again consistently outperforms the baseline strategy across the range of limb lengths. 
% maintains a stable cleaning percentage across various arm lengths and experiences a slight decrease with increasing leg length. In contrast, the baseline exhibits a slight decline as both the arm and leg lengths increase. 
These trends underscore the adaptability and consistent efficacy of the SkinGrip in contrast to the baseline end effector when evaluated on individuals with variations in limb size. % which may be less equipped to handle variations in limb size during the cleaning task.

% \begin{table}[h]
% \centering
% \caption{Demographics}
% \begin{tabular}{lcc}
% \hline
% Number of Participants & 12 \\
% Age (years) & $26.9 \pm 6.5$ & \\
% Weight (kg) & $73.3 \pm 9.3$ & \\
% Height (cm) & $176.6 \pm 8.2$ & \\
% Arm Diameter (cm) & $7.8 \pm 0.3$ \\
% Arm Length (cm) & $49.4 \pm 3.9$ \\
% Leg Diameter (cm) & $10.0 \pm 0.5$ \\
% Leg Length (cm) & $38.9 \pm 4.4$ \\

% \hline
% \end{tabular}
% \end{table}

\vspace{-0.2cm}
\subsection{Participant Responses}
\label{sec:response}

After each trial, the participants evaluated their experience using four $7$-point Likert scale questions: 1) \textit{`I felt safe during interaction with the robot'}, 2) \textit{`I felt comfortable with the robot interacting with me'}, 3) \textit{`The robot cleaned off my entire limb'}, and 4) \textit{`The robot cleaned my limb in a reasonable amount of time'}. Following the completion of all trials, participants indicated their preferred manipulator and the reasons for their choice. They selected one or more reasons from a set of the following options: \textit{`it felt safer'}, \textit{`it was more comfortable'}, \textit{`it cleaned off a larger surface area of my limb'}, \textit{`it took less time to clean my limb'}, \textit{`it is durable'}, \textit{`it fits well'}, \textit{`it is pain free'}, \textit{`it looks good'}, or provide their own reason in a blank space provided.

\begin{figure}[t]
      \centering
      \includegraphics[width = 1\columnwidth, trim={0cm 5cm 0cm 21cm}, clip]{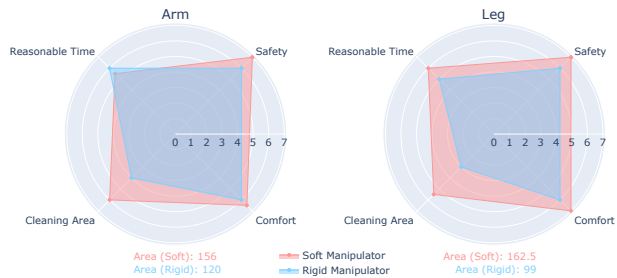}

      \caption{The median Likert responses for all participants for the soft and baseline manipulators for the arm and leg cleaning trials. The area of the radar charts is included for comparison. A larger surface area indicates higher Likert scores from participants across the four items.}
      \label{radarchart}

\end{figure}

\begin{figure}[t]
      \centering
      \includegraphics[width=0.5\textwidth, trim={0cm 14cm 0cm 15cm}, clip]{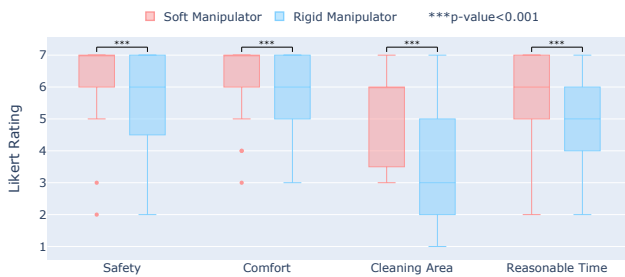}
      
      \caption{Boxplots showing participant responses to $4$ Likert items on \textit{Safety}, \textit{Comfort}, \textit{Cleaning Area}, and \textit{Reasonable Time} for the soft and baseline manipulators. Using a Wilcoxon Signed Rank Test, strong statistical significance is seen between the soft and baseline manipulator for all four items.}
      \label{boxplot}
      % \vspace{-0.3cm}
\end{figure}
% \vspace{-0.4cm}

Fig.~\ref{radarchart} shows the median responses of the Likert scale of the participants for both soft and baseline manipulators during arm and leg cleaning trials. 
% In arm trials, the SkinGrip scored higher in \textit{safety}, \textit{comfort}, and \textit{cleaning area} ($3$ out of $4$ categories), while in the \textit{reasonable time} category, the baseline end effector scored higher than the SkinGrip ($5.5$ for soft, $6$ for baseline). 
Across all arm trials, the SkinGrip outperformed in three of four categories: \textit{safety}, \textit{comfort}, and \textit{cleaning area}. Participants reported a slightly higher score for the baseline manipulator in terms of \textit{reasonable time} to complete the task (median of $5.5$ for soft, $6$ for baseline). 
Across leg trials, the proposed SkinGrip outperformed the baseline end effector in all categories. Additionally, as shown in Fig.~\ref{radarchart}, radar chart areas are provided for further comparison: $156$ units$^{2}$ for the SkinGrip and $120$ units$^{2}$ for the baseline manipulator in arm trials; $162.5$ units$^{2}$ for the SkinGrip and $99$ units$^{2}$ for the baseline manipulator in leg trials. This suggests a general preference for the SkinGrip by participants across key assessment factors. 
% This is likely because the lower leg is shorter than the arm (arm: $49.4(3.9)$ mean (SD) cm, leg: $38.9(4.4)$ cm), resulting in a faster cleaning time for the leg. Additionally, since the leg has a larger diameter compared to the arm (arm: $7.8(0.3)$ cm, leg: $10.0(0.5)$ cm), participants might have felt less pressure when the SkinGrip moves along the arm in comparison to the leg.

Fig.~\ref{boxplot} presents box plots of participant responses for the $4$ Likert items in all trials. For all items, the median response for the SkinGrip was higher than the baseline end effector across all trials. The most significant difference was in the \textit{`Cleaning Area'} item with the median response for the SkinGrip reported as $6$ (Agree) versus $3$ (Slightly Disagree) for the baseline end effector. The images in Fig.~\ref{fig:setup} (b) support these findings. Median ratings for safety and comfort were $7$ (Strongly Agree) for the SkinGrip and $6$ (Agree) for the baseline. A Wilcoxon signed-rank test (N = $48$) confirmed statistically significant differences between responses for the two manipulators for all items (p-values $<0.001$).

% Fig.~\ref{boxplot} shows boxplots for all participant responses to the $4$ Likert items after each trial for the soft and rigid manipulators. As seen, for all items, the median response for the SkinGrip was higher than the rigid manipulator across all trials. The largest difference was in the Cleaning Area item with the median response for the SkinGrip at a $6$ (Agree) in comparison to $3$ (Slightly Disagree) for the rigid manipulator. This finding is consistent with what we observed in the camera images, examples of which can be seen in Fig.~\ref{fig:setup}(b). Participant responses indicate that they felt safe and comfortable during interactions with the SkinGrip with median Likert ratings of $7$ (Strongly Agree) for the Likert Items on Safety and Comfort. Participants felt the same for the rigid manipulator but to a lesser degree with median Likert ratings of $6$ (Agree) for both items. By applying a Wilcoxon signed-rank test (N = $48$), we observe strong statistically significant differences between the soft and rigid manipulator for all $4$ Likert Items with all p-values $<0.001$. The signed-rank test is applied between the $48$ trials ($12$ participants, $2$ trials on the arm, $2$ trials on the leg) for the SkinGrip and the corresponding trials with the rigid manipulator. 

At the conclusion of the study, all participants ($12/12$) reported preferring the SkinGrip. The primary reported reason was that for the SkinGrip, \textit{`it cleaned off a larger surface area of my limb'}, with all $12$ participants reporting this. Additionally, $7$ participants reported \textit{`it was more comfortable'}, and $5$ reported \textit{`it felt safer'}.

\subsection{Discussion}
%In this work, we successfully developed a unique soft robotic manipulator for safe, comfortable, and effective bathing assistance. 
Through this research, we have illustrated the practicality and potential benefits of our soft robotic manipulator in enhancing the quality of bathing assistance, with an emphasis on safety, comfort, and effectiveness. However, our work builds on a few assumptions that could benefit from future advances. One key assumption is the expectation that an individual will be able to position their limb static, lifted off of a bed by a limb support. Second, we evaluated the proposed SkinGrip primarily on limb bathing scenarios. Although our SkinGrip has potential to clean other body parts, such as the back and torso, the control methods required for these tasks are still necessary. While the $1$-DOF soft fingers enable effective cleaning, future work could explore higher-DOF designs to further improve adaptability on complex body contours. An additional future development may include the design of computer vision techniques to track regions of a limb that have not been fully cleaned in order to perform a targeted cleaning maneuver around those regions.

% , our method presupposes that the user will position their limb on a holder and maintain a static pose for bathing assistance, which may not always be feasible in practical scenarios. Second, our study focuses mainly on the care of the limbs. Although our SkinGrip has potential capabilities to clean other body parts, such as the back and torso, the control methods required for these tasks have yet to be developed. Furthermore, our approach involves measuring the limb length and setting fixed values for the robot's movement along the limb, instead of autonomously detecting the start and end points. Future research could overcome this limitation by integrating an RGB-D camera to autonomously detect key points like the wrist and shoulder, enabling the robot to adjust its manipulator movements along the limb. This advancement would improve the functionality and applicability of the robotic system for a wider range of bathing assistance tasks.
% \fukang{Should we note the possibility that high skin moisture may affect the control method's performance due to the capacitance values' sensitivity to wet conditions? This point could impact the perceived practicality of our SkinGrip and might influence reviewers' opinions on our work's contributions.}

\section{Conclusion}
\label{sec:conclusion}

In this work, we introduced an innovative soft deformable robotic manipulator—SkinGrip, designed for high adaptability and conforming to the contours of human limbs.  By incorporating a capacitive servoing control scheme, the SkinGrip is able to detect contact and tightness of its soft fingers around the circumference of a human limb, while also effectively traversing the limb to perform robot-assisted bed bathing. We evaluate the SkinGrip and control strategy in a human experiment in which a mobile manipulator assists with cleaning off the arm and leg of 12 human participants. Our experiments highlight the SkinGrip and capacitive servoing control strategy's ability to clean and adapt to various limb sizes, including differences in diameter and length. Notably, the SkinGrip achieves an overall average of $88.8\%$ cleaning effectiveness on the arm and $81.4\%$ on the leg, significantly outperforming a baseline rigid end effector design. Participant-provided feedback reinforces this finding, in which we observe a statistically significant difference in reported preferences for a soft manipulation strategy in terms of safety, comfort, and thorough skin bathing capabilities.
% Additionally, user feedback indicates a statistically significant preference for the SkinGrip, particularly noting its safety, comfort, and thorough cleaning capabilities. Overall, this study underscores the versatility of the SkinGrip and its promising applications in robot-assisted bathing and other personal care tasks, enhancing human-robot interactions.

% there is possibility that the skin moisture may matter (impact capacitance 

% \section*{Acknowledgment}

% \small
% \textit{Text} 

\bibliographystyle{IEEEtran}
\bibliography{bibliography}

% \balance

\end{document}